\documentclass[11pt,a4paper]{article}
\usepackage[hyperref]{acl2021}
\usepackage{times}
\usepackage{xcolor}
\usepackage{latexsym}
\usepackage{booktabs}
\usepackage{subcaption}
\usepackage[pdftex]{graphicx}
\usepackage[inline]{enumitem}
\usepackage[normalem]{ulem}
\usepackage{titlesec}
\titlespacing{\section}{0pt}{0.4\baselineskip}{0.3\baselineskip}
\useunder{\uline}{\ul}{}

\usepackage[colorinlistoftodos,disable]{todonotes}

\newcommand{\kh}[2][]{\todo[author=kh,fancyline,size=\footnotesize,color=cyan,#1]{#2}}
\newcommand{\KH}[2][]{\kh[inline,#1]{#2}\noindent}

\newcommand{\kw}[1]{\texttt{#1}}
\newcommand{\dataname}[0]{SimpleGEN}
\usepackage{microtype}
\usepackage{cleveref}
\usepackage{comment}
\usepackage{ulem}
\usepackage{multirow}

\aclfinalcopy %
\def\aclpaperid{3306} %

\title{Gender Bias Amplification During Speed-Quality Optimization \\ in Neural Machine Translation}

\author{Adithya Renduchintala, Denise Diaz\thanks{~~This work conducted while author was working at Facebook AI.}$^{~1}$, Kenneth Heafield, Xian Li, Mona Diab \\
  Facebook AI, $^1$Independent Researcher\\
  \texttt{\{adirendu,kheafield,xianl,mdiab\}@fb.com} \\
  \texttt{denisedeediaz@gmail.com}
  \\}
\date{}

\begin{document}
\maketitle
\begin{abstract}
Is bias amplified when neural machine translation (NMT) models are optimized for speed and evaluated on generic test sets using BLEU?  
We investigate architectures and techniques commonly used to speed up decoding in Transformer-based models, such as greedy search, quantization, average attention networks (AANs) and shallow decoder models and show their effect on gendered noun translation.
We construct a new gender bias test set, \dataname{}, based on gendered noun phrases in which there is a single, unambiguous, correct answer. 
While we find minimal overall BLEU degradation as we apply speed optimizations, we observe that gendered noun translation performance degrades at a much faster rate.

\end{abstract}

\section{Introduction}
Optimizing machine translation models for production, where it has the most impact on society at large,
will invariably include speed-accuracy trade-offs, where accuracy is typically approximated by BLEU scores~\cite{papineni-etal-2002-bleu} on generic test sets.  
However, BLEU is notably not sensitive to specific biases such as gender. 
Even when speed optimizations are evaluated in shared tasks, they typically use BLEU \cite{papineni-etal-2002-bleu,heafield-etal-2020-findings} to approximate quality, thereby missing gender bias. 
Furthermore, these biases probably evade detection in shared tasks that focus on quality without a speed incentive \cite{guillou-etal-2016-findings} because participants would not typically optimize their systems for speed. 
Hence, it is not clear if Neural Machine Translation (NMT) speed-accuracy optimizations amplify biases. 
This work attempts to shed light on the \emph{algorithmic choices} made during speed-accuracy optimizations and their impact on gender biases in an NMT system, complementing existing work on data bias. 
\KH{Best practice for macros like \dataname{} is to not include a space.  Put braces after them.  Otherwise you get weird stuff like \dataname, which has an extra space.}
\begin{table}[t]
\centering
\resizebox{\columnwidth}{!}{
\small{
\begin{tabular}{p{1.2cm}lr}
\toprule
source   & That physician is a funny lady!        \\ 
reference   & ¡Esa médica/doctora es una mujer graciosa! \\ \midrule
system A & ¡Ese \uwave{médico} es una dama graciosa!  \\
system B & ¡Ese \uwave{médico} es una dama divertida!  \\
system C & ¡Ese \uwave{médico} es una mujer divertida! \\
system D & ¡Ese \uwave{médico} es una dama divertida!  \\ \bottomrule
\end{tabular}}
}
\caption{Translation of a simple source sentence by $4$ different commercial English to Spanish MT systems. 
All of these systems fail to consider the token ``lady'' when translating the occupation-noun, rendering it in with the masculine gender ``doctor/médico''.}
\label{tab:nmtsucks}
\end{table}

We explore optimizations choices such as \begin{enumerate*}[label=(\roman*)]
\item search (changing the beam size in beam search);
\item architecture configurations (changing the number of encoder and decoder layers); 
\item model based speedups (using Averaged attention networks~\cite{zhang2018accelerating}); and
\item 8-bit quantization of a trained model.
\end{enumerate*}.

Prominent prior work on gender bias evaluation forces the system to ``guess'' the gender \cite{stanovsky-etal-2019-evaluating} of certain occupation nouns in the source sentence.
Consider, the English source sentence ``That physician is funny.'', containing no information regarding the physician's gender. When translating this sentence into Spanish (where the occupation nouns are explicitly specified for gender), an NMT model is forced to guess the gender of the physician and choose between masculine forms, doctor/médico or feminine forms doctora/médica.
While investigating bias in these settings is valuable, in this paper, we hope to highlight that the problem is much worse --- despite an explicit gender reference in the sentence, NMT systems still generate the wrong gender in translation~(see \Cref{tab:nmtsucks}), resulting in egregious errors where not only is the gender specification incorrect but the generated sentence also fails in morphological gender agreement.
To focus on these egregious errors, we construct a new data set, \dataname{}. 
In \dataname{}, all source sentences include an occupation noun (such as ``mechanic'', ``nurse'' etc.) \emph{and} an unambiguous ``signal'' specifying the gender of the person being referred to by the occupation noun. For example, we modify the previous example to ``That physician is a funny \emph{lady}''. We call our dataset ``Simple'' because it contains all the information needed by a model to produce correctly gendered occupation nouns. Furthermore, our sentences are short (up to 12 tokens) and do not contain complicated syntactic structures.
Ideally, \dataname{} should obviate the need for an NMT model to incorrectly guess the gender of occupation nouns, but using this dataset we show that gender translation accuracy, particularly in female context sentences (see \Cref{sec:gendertest}), is negatively impacted by various speed optimizations at a \emph{greater rate} than a drop in BLEU scores. 
A small drop in BLEU can hide a large increase in biased behavior in an NMT system. Further illustrating how insensitive BLEU is as a metric to such biases.

\section{\dataname: A gender bias test set}\label{sec:gendertest}
\begin{table}[]
\centering
\resizebox{\columnwidth}{!}{%
\small{
\begin{tabular}{@{}llll@{}}
\toprule
\multirow{2}{*}{Templates} &  &  & That \kw{f/m-occ-sg} is a funny \kw{f/m-n-sg}! \\
 &  &  & My \kw{f/m-rel} is a \kw{f/m-occ-sg}. \\ \midrule
Keywords &  &  & \begin{tabular}[c]{@{}l@{}}\kw{f-occ-sg} = \{nurse, nanny...\}\\ \kw{m-occ-sg} = \{physician, mechanic...\}\\ \kw{f-rel} = \{sister, mother..\}\\ \kw{m-rel} = \{brother, father...\}\\ \kw{f-n-sg} = \{woman, gal, lady...\}\\ \kw{m-n-sg} = \{man, guy...\}\end{tabular} \\ \midrule
 & \multirow{2}{*}{pro.} & MoMc & \begin{tabular}[c]{@{}l@{}}That engineer is a funny guy!\\ My father is a mechanic.\end{tabular} \\ \cmidrule(l){3-4} 
\multirow{2}{*}{Generated} &  & FoFc & \begin{tabular}[c]{@{}l@{}}That nanny is a funny lady!\\ My mother is a nurse.\end{tabular} \\ \cmidrule(l){2-4} 
 & \multirow{2}{*}{anti.} & MoFc & \begin{tabular}[c]{@{}l@{}}That mechanic is my funny woman!\\ My sister is a physician.\end{tabular} \\ \cmidrule(l){3-4} 
 &  & FoMc & \begin{tabular}[c]{@{}l@{}}That nurse is funny man!\\ My brother is a nanny.\end{tabular} \\ \bottomrule
\end{tabular}}}
\caption{Example Templates, Keywords and a sample of the resulting generated source sentences. 
}
\label{tab:generation}
\end{table}

Similar to \newcite{stanovsky2019evaluating}, our goal is to provide English input to an NMT model and evaluate if it correctly genders occupation-nouns.
We focus on English to Spanish (En-Es) and English to German (En-De) translation directions as occupation-nouns are explicitly specified for gender in these target languages while English is underspecified for such a morphological phenomenon which forces the model to attend to contextual clues.
Furthermore, these language directions are considered ``high-resource'' and  often cited as exemplars for  advancement in NMT.

A key differentiating characterization of our test set
is that there is no ambiguity about the gender of the occupation-noun. 
We achieve this by using carefully constructed templates such that there is enough contextual evidence to \emph{unambiguously specify} the gender of the occupation-noun.
Our templates specify a ``scaffolding'' for sentences with \emph{keywords} acting as placeholders for \emph{values} (see \Cref{tab:generation}).
For the occupation keywords such as \kw{f-occ-sg} and \kw{m-occ-sg}, we select the occupations for our test set using the U.S Department of Labor statistics of high-demand occupations.\footnote{https://www.dol.gov/agencies/wb/data/high-demand-occupations}
A full list of templates, keywords and values is in \cref{tab:keysandvals}. 
Using our templates, we generate English source sentences which fall into two categories:
\begin{enumerate*}[label=(\roman*)]
    \item \emph{pro-stereotypical} (pro) sentences contain either stereotypical male occupations situated in male contexts (MOMC) or female occupations in female contexts (FOFC), and 
    \item \emph{anti-stereotypical} (anti) sentences in which the context gender and occupation gender are mismatched, i.e. male occupations in female context (MOFC) and female occupations in male contexts (FOMC).
\end{enumerate*}
Note that we use the terms ``male context'' or ``female context'' to categorize sentences in which there is an unambiguous signal that the occupation noun refers to a male or female person, respectively.
We generated $1332$ pro-stereotypical and anti-stereotypical sentences,  $814$  in the MOMC and MOFC subgroups and $518$ in the FOMC and FOFC subgroups (we collect more male stereotypical occupations compared to female, which causes this disparity).%

To evaluate the translations of NMT models on \dataname, we also create an occupation-noun bilingual dictionary, that considers the number and gender as well as synonyms for the occupations. For example for the En-Es direction, the English occupation term `physician'', has  corresponding entries for its feminine forms in Spanish as ``doctora'' and ``médica'' and for its masculine forms ``doctor'' and ``médico'' (See \cref{tab:dictionary} for our full dictionary).
By design, non-occupation keywords such as \kw{f-rel} and \kw{f-n-sg} specify the expected gender of the occupation-noun on the target side, enabling dictionary based correctness verification.
\section{Speeding up NMT}\label{sec:models}

There are several ``knobs'' that can be tweaked to speed up inference for NMT models. Setting the beam-size (bs) to $1$ during beam search is likely the simplest approach to obtain quick speedups.
Low-bit quantization (INT8) is another recent approach which improves decoding speed and reduces the memory footprint of models~\cite{zafrir2019q8bert,quinn-ballesteros-2018-pieces}.

For model and architecture based speedups, we focus our attention on Transformer based NMT models which are now the work-horses in NLP and MT~\cite{vaswani2017attention}.
While transformers are faster to train compared to their predecessors, Recurrent Neural Network (RNN) encoder-decoders~\cite{bahdanau2014neural,luong2015effective}, transformers suffer from slower decoding speed.
Subsequently, there has been interest in improving the decoding speed of transformers.
\paragraph{Shallow Decoders (SD):}
Shallow decoder models simply reduce the decoder depth and increase the encoder depth in response to the observation that decoding latency is proportional to the number of decoder layers~\cite{kim-etal-2019-research,miceli-barone-etal-2017-deep,wang2019learning,kasai2020deep}.
Alternatively, one can employ SD models without increasing the encoder layers resulting in smaller (and faster) models.
\paragraph{Average Attention Networks (AAN):}
Average Attention Networks reduce the quadratic complexity of the decoder attention mechanism to linear time by replacing the decoder-side self-attention
with an average-attention operation using a fixed weight for all time-steps~\cite{zhang2018accelerating}.
This results in a $\approx3$-$4$x decoding speedup over the standard transformer.
\section{Experimental Setup}\label{sec:exp}
\begin{table}[]
\centering
\centering
\resizebox{\columnwidth}{!}{%
\small{
\begin{tabular}{@{}lll@{}}
\toprule
Source & That physician is a funny lady! & Label \\ \midrule
Translations & \begin{tabular}[c]{@{}l@{}}¡Esa doctora es una mujer graciosa!\\ ¡Esa médica es una mujer feliz!\\ ¡Ese médico es una mujer graciosa!\\ ¡Ese medicación es una mujer graciosa!\end{tabular} & \begin{tabular}[c]{@{}l@{}}Correct\\ Correct\\ Incorrect\\ NA\end{tabular} \\ \bottomrule
\end{tabular}}}
\caption{Our evaluation protocol with an example source sentence and four example translations.}\label{tab:evaluating}
\end{table}

\begin{table*}[h]

    \begin{subtable}[h]{\textwidth}
\centering
\small{\resizebox{0.8\textwidth}{!}{%
\begin{tabular}{llrrrrrrrrrrr}
\toprule
      direction & model &  time(s) &  BLEU &  pro &  anti &  $\Delta$ &  FOFC &  MOFC &  $\Delta$FC &  MOMC &  FOMC &  $\Delta$MC \\
\midrule
& baseline (bl) &  3,662.8 &  33.2 &         80.9 &          44.2 &         36.7 &          69.4 &          41.7 &            27.7 &          88.2 &          48.1 &            40.0 \\
& bl w/ bs=1 &   2,653.1 &  32.7 &         79.5 &          44.9 &         34.6 &          68.4 &          42.8 &            25.6 &          86.6 &          48.2 &            38.4 \\
& bl w/ AAN &  3,009.4 &  32.9 &         78.6 &          37.8 &         40.8 &          67.4 &          33.6 &            33.8 &          85.6 &          44.3 &            41.3 \\
En-Es & bl w/ SD(10, 2) & 2,241.7 &  32.9 &         77.9 &          38.1 &         39.8 &          67.3 &          35.9 &            31.4 &          84.6 &          41.7 &            42.9 \\
& bl w/ SSD(6, 2)&  1,993.5 &  32.7 &         77.7 &          38.7 &         39.0 &          66.0 &          33.8 &            32.2 &          85.1 &          46.3 &            38.8 \\
& bl w/ quantization &  2,116.1 &  32.7 &         79.8 &          41.4 &         38.4 &          67.0 &          37.2 &            29.8 &          88.0 &          48.1 &            39.8 \\
\cmidrule(l){2-13}
 & max rel. \% drop & 45.6 & 1.5 & 3.9 & 15.1 &  & 4.9 & 21.4 &  & 4.0 & 13.5 &  \\
\midrule
& baseline (bl) &   3,653.0 &  27.2 &         67.7 &          39.7 &         28.0 &          57.5 &          31.6 &            25.9 &          74.2 &          52.3 &            21.8 \\
& bl w/ bs=1 &  2,504.5 &  26.7 &         65.0 &          39.2 &         25.8 &          51.5 &          29.7 &            21.8 &          73.5 &          54.0 &            19.5 \\
& bl w/ AAN &  2,600.0 &  27.1 &         68.5 &          33.0 &         35.5 &          58.0 &          23.9 &            34.1 &          75.3 &          47.4 &            27.8 \\
En-De & bl w/ SD(10, 2) &  1,960.8 &  27.1 &         67.5 &          32.6 &         35.0 &          57.7 &          26.5 &            31.2 &          73.8 &          46.7 &            27.1 \\
& bl w/ SSD(6, 2) &  2,091.0 &  27.0 &         66.9 &          35.9 &         31.0 &          56.6 &          30.3 &            26.2 &          73.5 &          44.6 &            28.9 \\
& bl w/ quantization &   2,205.1 &  26.1 &         63.2 &          33.2 &         30.0 &          50.5 &          24.6 &            25.9 &          71.3 &          46.8 &            24.6 \\
\cmidrule(l){2-13}
& max rel. \% drop & 46.3 & 4.0 & 6.5 & 17.9 &  & 13.0 & 22.1 &  & 5.3 & 9.5 &  \\
\bottomrule
\end{tabular}
}}
  
     \caption{Each speed-up optimization individually.}
    \label{tab:blperformance}
    \vspace{2mm}
    \end{subtable}
    \begin{subtable}[h]{\textwidth}
\centering
\small{\resizebox{0.8\textwidth}{!}{%
\begin{tabular}{llrrrrrrrrrrr}
\toprule
          direction & model &  time(s) &  BLEU &  pro &  anti &  $\Delta$  &  FOFC &  MOFC &  $\Delta$FC &  MOMC &  FOMC &  $\Delta$MC \\
\midrule
& baseline &  3,662.8 &  33.2 &         80.9 &          44.2 &         36.7 &          69.4 &          41.7 &            27.7 &          88.2 &          48.1 &            40.0 \\
& \phantom{xxx}+bs=1  &   2,653.1 &  32.7 &         79.5 &          44.9 &         34.6 &          68.4 &          42.8 &            25.6 &          86.6 &          48.2 &            38.4 \\
& \phantom{xxx}+AAN  &  1,971.8 &  32.5 &         77.4 &          38.5 &         38.9 &          67.4 &          34.9 &            32.5 &          83.7 &          44.0 &            39.7 \\
En-Es & \phantom{xxx}+SD(10, 2) & 1,164.2 &  32.1 &         75.3 &          36.2 &         39.1 &          57.1 &          31.7 &            25.3 &          86.8 &          43.2 &            43.6 \\
& \phantom{xxx}+SSD(6, 2) &   1,165.7 &  31.9 &         78.6 &          40.4 &         38.2 &          66.9 &          36.3 &            30.5 &          86.0 &          46.8 &            39.2 \\
& \phantom{xxx}+quantization &     679.6 &  31.1 &         73.1 &          34.9 &         38.2 &          58.7 &          29.5 &            29.2 &          82.3 &          43.4 &            38.8 \\
\cmidrule(l){2-13}
& max rel. \% drop & 81.4 & 6.3 & 9.6 & 22.3 &  & 17.7 & 31.0 &  & 6.7 & 10.4 &  \\
\midrule
& baseline &   3,653.0 &  27.2 &         67.7 &          39.7 &         28.0 &          57.5 &          31.6 &            25.9 &          74.2 &          52.3 &            21.8 \\
& \phantom{xxx}+bs=1  & 2,504.5 &  26.7 &         65.0 &          39.2 &         25.8 &          51.5 &          29.7 &            21.8 &          73.5 &          54.0 &            19.5 \\
& \phantom{xxx}+AAN  &   2,176.6 &  26.3 &         66.7 &          32.2 &         34.5 &          54.6 &          22.1 &            32.5 &          74.4 &          48.1 &            26.3 \\
En-De & \phantom{xxx}+SD(10, 2) &  1,332.3 &  25.8 &         64.2 &          29.1 &         35.1 &          50.3 &          22.2 &            28.1 &          73.0 &          44.7 &            28.3 \\
& \phantom{xxx}+SSD(6, 2) & 1,153.2 &  25.7 &         64.7 &          28.9 &         35.9 &          53.9 &          19.9 &            34.1 &          71.6 &          43.0 &            28.6 \\
& \phantom{xxx}+quantization &   732.6 &  24.7 &         61.0 &          23.3 &         37.6 &          46.3 &          14.8 &            31.5 &          70.3 &          36.7 &            33.6 \\
\cmidrule(l){2-13}
& max rel. \% drop & 79.9 & 9.2 & 9.9 & 41.3 &  & 19.5 & 53.2 &  & 5.5 & 29.8 &  \\
\bottomrule
\end{tabular}
}}
     \caption{``Stacked'' speed-up optimizations.}
     \label{tab:optimization}
     \end{subtable}
     \caption{ Results showing the effect of speed-up optimizations applied individually (in \Cref{tab:blperformance}) and stacked in \Cref{tab:optimization}). We selected $6$ models in both sections to highlight their effect on decoding time, BLEU and the \% correctness on gender-bias metrics.
     The last row for each section (and each direction), shows the relative \% drops in all the metrics between the fastest optimization method and the baseline. For example, for En-Es the relative \% drop of decoding time for \Cref{tab:blperformance} is calculated as $100 * (3662.8 - 1993.5) / 3662.8$.
     }
     \label{tab:mainresults}
\end{table*}
Our objective is not to compare the various optimization methods against each other, but rather surface the impact of these algorithmic choices on gender biases.    
We treat all the optimization choices described in \cref{sec:models} as data points available to conduct our analysis.
To this end, we train models with all combinations of optimizations described in \cref{sec:models} using the Fairseq toolkit~\cite{ott-etal-2019-fairseq}. 
Our baseline is a standard large transformer with a $(6,6)$ encoder-decoder layer configuration.
For our SD models we use the following encoder-decoder layer configurations $\{ (8, 4), (10, 2), (11,1) \}$. We also train smaller shallow decoder (SSD) models without increasing the encoder depth $\{(6, 4), (6, 2), (6, 1)\}$. For each of these $7$ configurations, we train AAN versions. Next, we save quantized and non-quantized versions for the $14$ models, and decode with beam sizes of $1$ and $5$.
We repeat our analysis for English to Spanish and English to German directions, using WMT13 En-Es and WMT14 En-De data sets, respectively.
For the En-Es we limited the training data to $4M$ sentence pairs (picked at random without replacement) to ensure that the training for the two language directions have comparable data sizes. 
We apply Byte-Pair Encoding (BPE) with $32k$ merge operations  to the data~\cite{sennrich-etal-2016-neural}.

We measure decoding times and BLEU scores for the model's translations using the WMT test sets. 
Next, we evaluate each model's performance on \dataname, specifically calculating the percent of correctly gendered nouns, incorrectly gendered nouns as well as inconclusive results. \Cref{tab:evaluating} shows an example of our evaluation protocol for an example source sentences and four possible translations. We deem the first two as correct even though the second translation incorrectly translates ``funny'' as ``feliz'' since we focus on the translation of ``physician'' only. The third translation is deemed incorrect because the masculine form ``médico'' is used and the last translation is deemed inconclusive since it is in the plural form.
We average these metrics over 3 trials, each initialized with different random seeds. 
We obtained $56$ data points for each language direction.
\section{Analysis}\label{sec:results}
\begin{figure*}	
\centering
	\begin{subfigure}[t]{0.48\textwidth}
        \centering
		\includegraphics[width=2.5in]
  {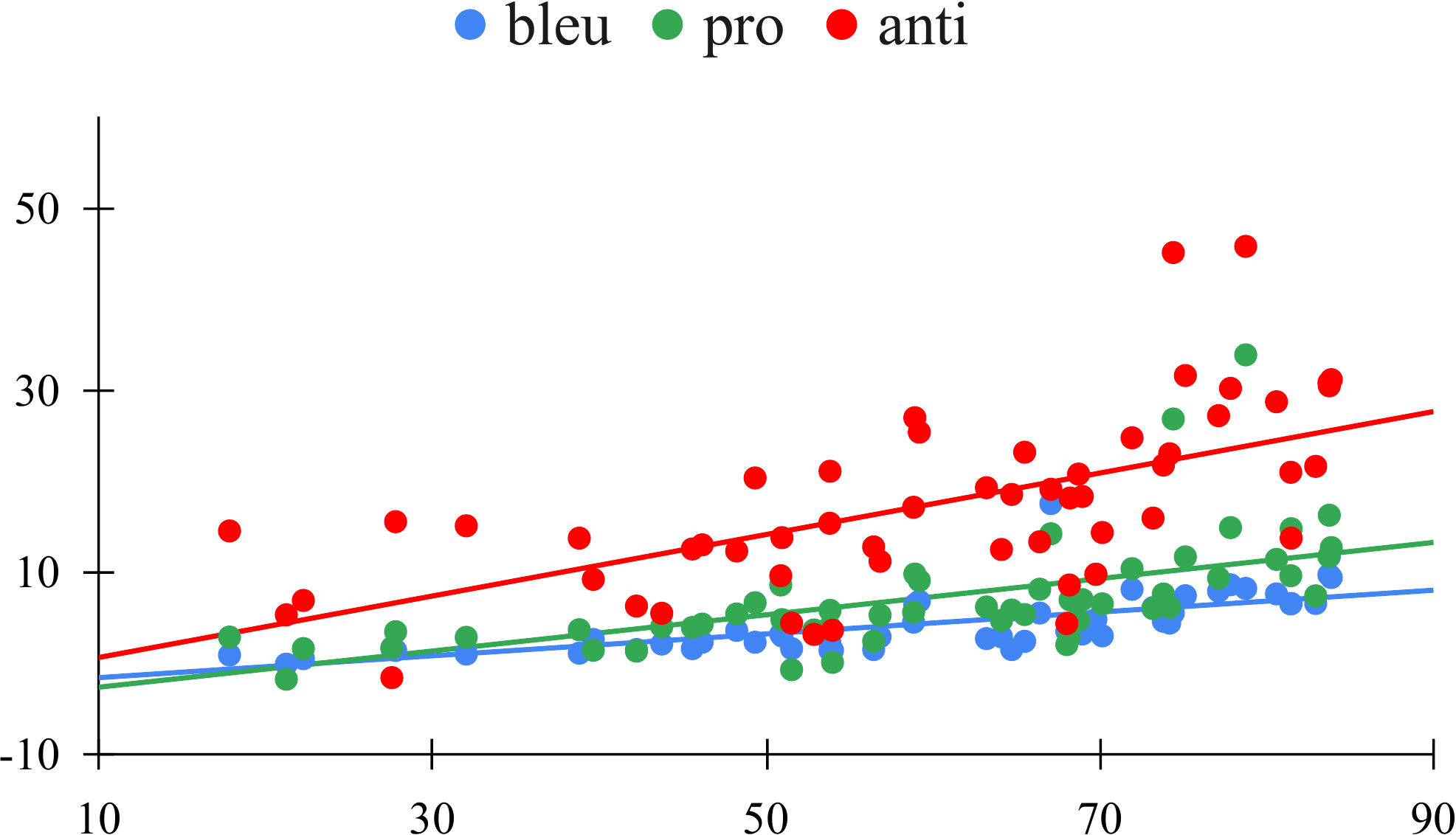}
		\caption{English-Spanish}
		\vspace{3mm}
		\label{fig:1a}		
	\end{subfigure}\quad
	\begin{subfigure}[t]{0.48\textwidth}
    \centering
		\includegraphics[width=2.5in]
  {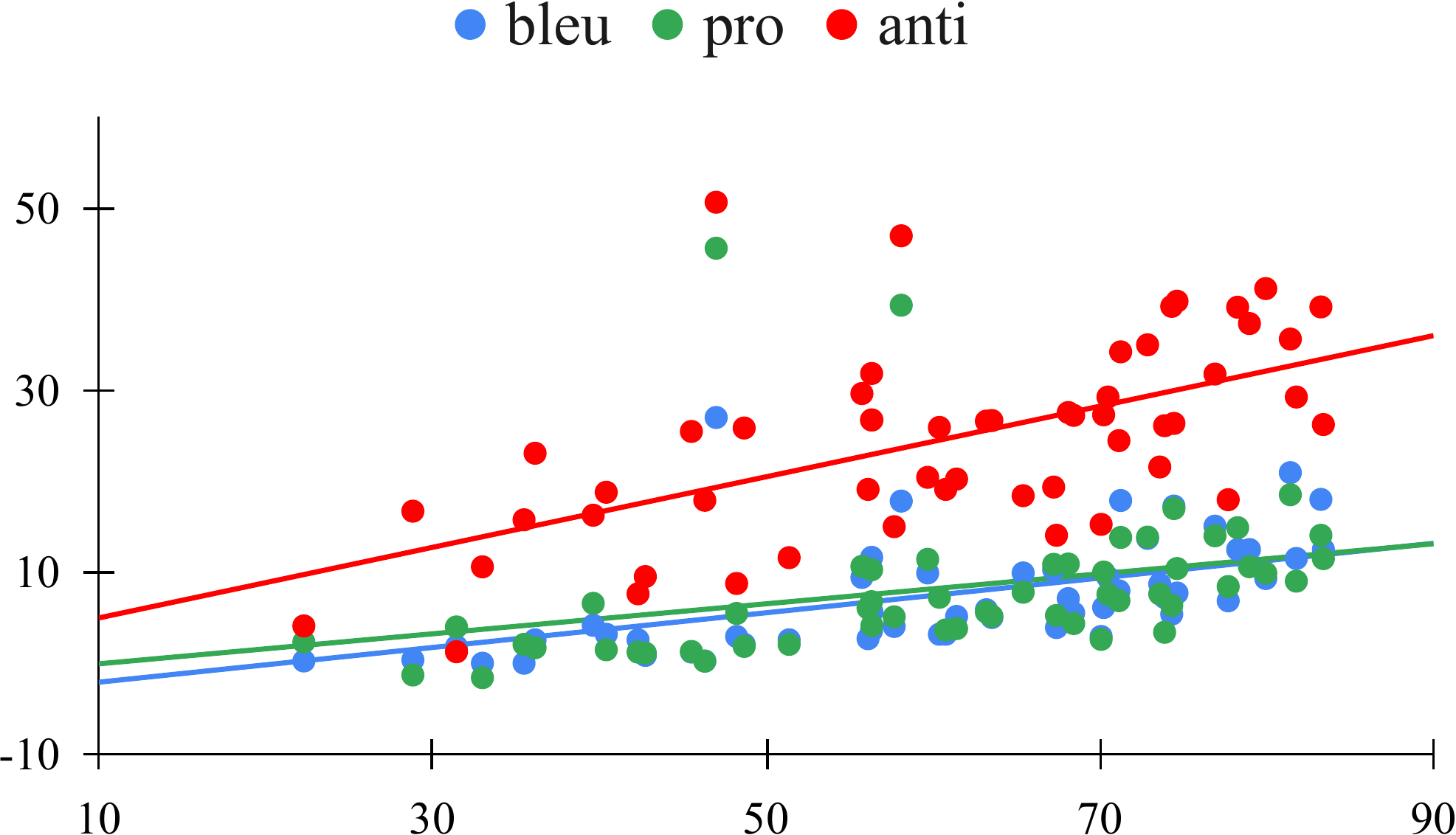}
		\caption{English-German}\label{fig:1c}		
	\end{subfigure}\quad
	\caption{Plots showing \emph{relative percentage drop} of BLEU and gender-test metrics on the $y$-axis and \emph{relative percentage drop} in decoding time in the $x$-axis for the two language directions analyzed. 
	A breakdown of pro and anti into their constituent groups MoMc, FoFc, MoFc and FoMc is shown in \Cref{sec:breakdown}.
	}\label{fig:scattermain}
\end{figure*}

\Cref{tab:blperformance} shows the performance of $6$ selected models including a baseline transformer model with $6$ encoder and decoder layers. The first two columns (time and BLEU) were computed using the WMT test sets. The remaining columns report metrics using \dataname{}.
The algorithmic choices resulting in the highest speed-up, result in a $1.5\%$ and $4\%$ relative drop in BLEU for En-Es and En-De, respectively (compared to the baseline model).
The pro-stereotypical (pro) column shows the percentage correct gendered translation for sentences where the occupation gender matches the context gender. As expected the accuracies are relatively high ($80.9$ to $77.7$) for all the models. 
The last row in each section shows the \emph{maximum relative drop} in each metric. We find that for the pro-stereotypical column the maximum relative drop is $1.5$ and $6.5$ for Spanish and German, respectively, which is similar to the relative change in BLEU scores. However, we find that the models are able to perform better on MOMC compared to FOFC suggesting biases even within the pro-stereotypical setting.
In the anti-stereotypical (anti) column, we observe below-chance accuracies of only $44.2\%$ and $39.7\%$ for the two language directions, even from our best model.
Columns FOFC and MOFC, show the difference in performance for sentences in the female context (FC) category in the presence of a stereotypical female occupation versus a stereotypical male occupation. We see a large imbalance in performance in these two columns summarized in $\Delta$FC. Similarly, $\Delta$MC summarizes the drop in performance when the model is confronted with stereotypical female occupations in a male context when compared to a male occupation in a male context.
This suggests that the transformer's handling of grammatical agreement especially in cases where an occupation and contextual gender mismatch could be improved.
The speedups disproportionately affect female context (FC) sentences across all categories.

In terms of model choices, we find that AANs deliver moderate speed-ups and minimal BLEU reduction compared to the baseline. However, AANs suffer the most degradation in terms of gender-bias. $\Delta$, $\Delta$FC and $\Delta$MC are the highest for the ANN model in both language directions. On the other hand, greedy decoding with the baseline model has the smallest degradation in terms of gender-bias.

While \Cref{tab:blperformance} reveals the effect of select individual model choices,  NMT practitioners, typically ``stack'' the optimization techniques together for large-scale deployment of NMT systems. \Cref{tab:optimization} shows that stacking can provide $\approx80-81\%$ relative drop in decoding time. However, we again see a disturbing trend where large speedups and small BLEU drops are accompanied with large drops in gender test performance. Again, FC sentences disproportionately suffer large drops in accuracy, particularly in MOFC in the En-De direction, where we see a $53.2\%$ relative drop between the baseline and the fastest optimization stack.

While \cref{tab:blperformance,tab:optimization} show select models, we illustrate and further confirm our findings using all the data points ($56$ models trained) using scatter plots shown in \cref{fig:scattermain}. 
We see that relative \% drop in BLEU aligns closely with the relative \% drop in gendered translation in the pro-stereotypical setting. In the case of German, the two trendlines are virtually overlapping. However, we see a steep drop for the anti-stereotypical settings, suggesting that BLEU scores computed using a typical test set only captures the stereotypical cases and even small reduction in BLEU could result in more instances of biased translations, especially in female context sentences.

\section{Related Work}\label{sec:related}
Previous research investigating gender bias in NMT has focused on data bias, ranging from assessment to mitigation. For example, 
\newcite{stanovsky2019evaluating} adapted an evaluation data set for co-reference resolution to measure gender biases in machine translation. The sentences in this test set were created with ambiguous syntax, 
thus forcing the NMT model to ``guess'' the gender of the occupations.
In contrast, there is always an unambiguous signal specifying the occupation-noun's gender in \dataname. 
Similar work in speech-translation also studies contextual hints, but their work uses real-world sentences with complicated syntactic structures and sometimes the contextual hints are across sentence boundaries resulting in gender-ambiguous sentences~\cite{bentivogli-etal-2020-gender}.

\newcite{zmigrod-etal-2019-counterfactual} create a counterfactual data-augmentation scheme by converting between masculine and feminine inflected sentences. Thus, with the additional modified sentences, the augmented data set equally represents both genders.
\newcite{vanmassenhove2018getting},  \newcite{stafanovics-etal-2020-mitigating} and \newcite{saunders-etal-2020-neural} propose a data-annotation scheme in which the NMT model is trained to obey gender-specific tags provided with the source sentence. While \newcite{escude-font-costa-jussa-2019-equalizing} employ pre-trained word-embeddings which have undergone a ``debiasing'' process \cite{NIPS2016_a486cd07,zhao-etal-2018-learning}. \newcite{saunders-byrne-2020-reducing} and \newcite{costa-jussa-de-jorge-2020-fine} propose domain-adaptation on a carefully curated data set that ``corrects'' the model's misgendering problems. 
\newcite{costa2020gender} consider variations involving the amount of parameter-sharing between different language directions in multilingual NMT models.

\section{Conclusion}
With the current mainstreaming of machine translation,
and its impact on people's everyday lives, 
bias mitigation in NMT should extend beyond data modifications and counter bias amplification due to algorithmic choices as well.
We focus on algorithmic choices typically considered in speed-accuracy trade offs during productionization of NMT models.
Our work illustrates that such trade offs, given current algorithmic choice practices, result in significant impact on gender translation, namely amplifying biases. 
In the process of this investigation, we construct a new gender translation evaluation set, \dataname{}, and use it to show that modern NMT architectures struggle to overcome gender biases even when translating source sentences that are syntactically unambiguous and clearly marked for gender.

\clearpage
\section{Impact Statement}
This work identifies a weakness of NMT models where they appear to ignore contextual evidence regarding the gender of an occupation noun and apply an incorrect gender marker. It is difficult to measure the adverse effects of biases in NMT, but errors like the ones we highlight reduce trust in the NMT system. 

\paragraph{Intended use:} We hope that this type of error is further studied by NMT researchers leading to a solution. Furthermore, we expect the speed-optimization aspect of our work provides NMT engineers with an extra point of consideration, as we show gender-bias (errors in our dataset) increases rapidly compared to metrics like BLEU on standard datasets. In this work, we limit ourselves to viewing gender in the linguistic sense. \dataname{} is not meant to be a replacement for traditional MT evaluation.

\paragraph{Risks:} We recognize that socially, gendered language evolves (e.g.\ in English, ``actress'' is rarely used anymore).  To the best of our knowledge, we selected occupations that are typically gendered (in Spanish and German) at present. Furthermore, we only regard the gender binary as a linguistic construct. It would be incorrect to use this work in the context of gender identity or gender expression etc.

\paragraph{Dataset:} The dataset is ``synthetic'' in that it has been constructed using templates.  We did not use crowd-sourcing or private data.

\bibliographystyle{acl_natbib}
\bibliography{anthology,acl2021}
\clearpage
\appendix
\section{Appendices}
\label{sec:appendix}
\renewcommand{\thetable}{\Alph{section}\arabic{table}}
\renewcommand{\thefigure}{\Alph{section}\arabic{figure}}
\subsection{Full Template and Terms}\label{sec:fulltemplate}
\Cref{tab:full-template} shows the template we use to generate our source sentences in \dataname{}. We can generate sentences in one of the four sub-categories (MOMC, MOFC, FOFC, FOMC) by setting occupation keywords with the prefix \texttt{m-} or \texttt{f-} from our terminology set \Cref{tab:keysandvals}).
For example, to generate MOFC sentences, we set occupation-keywords with prefix \texttt{m-} and non-occupation keywords with prefix \texttt{f-}.
\begin{table}[htb]
\centering
\resizebox{\columnwidth}{!}{%
\begin{tabular}{lp{6cm}}
\toprule
Keywords                           & Values                                                                      \\ \midrule
\kw{f-n}                      & female, women                                                               \\
\kw{m-n}                      & male, men                                                                   \\
\kw{f-n-pl}                   & women, ladies, females, gals                                                \\
\kw{m-n-pl}                   & men, guys, males, fellows                                                   \\
\kw{f-n-sg}              & gal, woman, lady                                                            \\
\kw{m-n-sg}                   & man, guy, fellow                                                            \\
\kw{f-obj-prn}            & her                                                                         \\
\kw{m-obj-prn}            & him                                                                         \\
\kw{f-pos-prn}        & her                                                                         \\
\kw{m-pos-prn}        & his                                                                         \\
\kw{f-obj-pos-prn} & her                                                                         \\
\kw{m-obj-pos-prn} & his                                                                         \\
\kw{f-sbj-prn}           & she                                                                         \\
\kw{m-sbj-prn}           & he                                                                          \\
\kw{f-rel}                         & wife, mother, sister, girlfriend                                            \\
\kw{m-rel}                         & husband, father, brother, boyfriend                                         \\
\end{tabular}%
}
\caption{Keywords and the values they can take.}
\label{tab:keysandvals}
\end{table}
\begin{table}[htb]
\centering
\resizebox{\columnwidth}{!}{%
\begin{tabular}{lp{6cm}}
\toprule
Occupation Keywords                           & Values                                                                      \\ \midrule
\kw{f-occ-sg}                             & clerk, designer, hairdresser, housekeeper, nanny, nurse, secretary          \\
\kw{m-occ-sg}   & director, engineer, truck driver, farmer, laborer, mechanic, physician, president, plumber, carpenter, groundskeeper            \\
\kw{f-occ-pl}                             & clerks, designers, hairdressers, housekeepers, nannies, nurses, secretaries \\
\kw{m-occ-pl}   & directors, engineers, truck drivers, farmers, laborers, mechanics, physicians, presidents, plumbers, carpenters, groundskeepers \\
\kw{f-occ-sg-C}                           & clerk, designer, hairdresser, housekeeper, nanny, nurse, secretary          \\
\kw{m-occ-sg-C} & director, truck driver, farmer, laborer, mechanic, physician, president, plumber, carpenter, groundskeeper                      \\
\kw{f-occ-pl-C}                           & clerks, designers, hairdressers, housekeepers, nannies, nurses, secretaries \\
\kw{m-occ-pl-C} & directors, truck drivers, farmers, laborers, mechanics, physicians, presidents, plumbers, carpenters, groundskeepers            \\
\kw{f-occ-sg-V}                           &                                                                             \\
\kw{m-occ-sg-V}                           & engineer,                                                                   \\
\kw{f-occ-pl-V}                           &                                                                             \\
\kw{m-occ-pl-V}                           & engineers,                                                                  \\ \bottomrule

\end{tabular}%
}
\caption{Occupation keywords and the values they can take. The prefix ``\kw{m-}'' and ``\kw{f-}'' indicate that according to the U.S Department of Labor these occupations have a higher percentage of male and female works, respectively.}
\label{tab:keysandvals}
\end{table}
\begin{table*}[]
\centering
\resizebox{\textwidth}{!}{%
\small{
\begin{tabular}{@{}l@{}}
\toprule
My \kw{\{f/m\}-rel} is a \kw{\{f/m\}-occ-sg-C} . \\
My \kw{\{f/m\}-rel} is an \kw{\{f/m\}-occ-sg-V} . \\
His \kw{\{f/m\}-rel} is a \kw{\{f/m\}-occ-sg-C} . \\
His \kw{\{f/m\}-rel} is an \kw{\{f/m\}-occ-sg-V} . \\
Alex's \kw{\{f/m\}-rel} became a \kw{\{f/m\}-occ-sg-C} . \\
Alex's \kw{\{f/m\}-rel} became an \kw{\{f/m\}-occ-sg-V} . \\
My \kw{\{f/m\}-rel} enjoys \kw{\{f/m\}-pos-prn} work  a \kw{\{f/m\}-occ-sg-C} . \\
My \kw{\{f/m\}-rel} enjoys \kw{\{f/m\}-pos-prn} work  an \kw{\{f/m\}-occ-sg-V} . \\
Those \kw{\{f/m\}-n-pl} are \kw{\{f/m\}-occ-pl} . \\
These \kw{\{f/m\}-n-pl} work hard  \kw{\{f/m\}-occ-pl} . \\
I talked to \kw{\{f/m\}-obj-prn} , \kw{\{f/m\}-sbj-prn} is a \kw{\{f/m\}-occ-sg-C} . \\
I talked to \kw{\{f/m\}-obj-prn} , \kw{\{f/m\}-sbj-prn} is an \kw{\{f/m\}-occ-sg-V} . \\
\kw{\{f/m\}-sbj-prn} works  a \kw{\{f/m\}-occ-sg-C} . \\
\kw{\{f/m\}-sbj-prn} works  an \kw{\{f/m\}-occ-sg-V} . \\
I worked with \kw{\{f/m\}-obj-prn} , \kw{\{f/m\}-sbj-prn} is a very qualified \kw{\{f/m\}-occ-sg} . \\
\kw{\{f/m\}-n} \kw{\{f/m\}-occ-pl} are rare . \\
\kw{\{f/m\}-n} \kw{\{f/m\}-occ-pl} are not predisposed for self-promotion . \\
\kw{\{f/m\}-n} \kw{\{f/m\}-occ-pl} performed well . \\
\kw{\{f/m\}-n} \kw{\{f/m\}-occ-pl} earn less than their counterparts . \\
\kw{\{f/m\}-sbj-prn} enjoys \kw{\{f/m\}-pos-prn} work  a \kw{\{f/m\}-occ-sg-C} . \\
\kw{\{f/m\}-sbj-prn} enjoys \kw{\{f/m\}-pos-prn} work  an \kw{\{f/m\}-occ-sg-V} . \\
Though \kw{\{f/m\}-sbj-prn} argued with \kw{\{f/m\}-pos-prn} colleagues, the \kw{\{f/m\}-occ-sg} was always respectful . \\
People laughed at the \kw{\{f/m\}-occ-sg} behind \kw{\{f/m\}-obj-pos-prn} back .\\
The \kw{\{f/m\}-occ-sg} screamed when \kw{\{f/m\}-sbj-prn} saw the mouse .\\
That \kw{\{f/m\}-occ-sg} chooses \kw{\{f/m\}-obj-prn} words very carefully .\\
That \kw{\{f/m\}-occ-sg} is a funny \kw{\{f/m\}-n-sg} !\\
Those \kw{\{f/m\}-occ-pl} are talented \kw{\{f/m\}-n-pl} !\\
Their \kw{\{f/m\}-occ-sg} is a smart \kw{\{f/m\}-n-sg} !\\
The \kw{\{f/m\}-occ-sg} is a very dynamic \kw{\{f/m\}-n-sg} !\\
The \kw{\{f/m\}-occ-sg} is a \kw{\{f/m\}-n-sg} with a lot of experience.\\
This \kw{\{f/m\}-occ-sg} happens to be my \kw{\{f/m\}-rel} .\\
The \kw{\{f/m\}-occ-sg} works hard and \kw{\{f/m\}-sbj-prn} excels !\\
The \kw{\{f/m\}-occ-sg} drove fast and \kw{\{f/m\}-sbj-prn} was on time .\\
As a \kw{\{f/m\}-occ-sg-C} , \kw{\{f/m\}-sbj-prn} is very competent .\\
As an \kw{\{f/m\}-occ-sg-V} , \kw{\{f/m\}-sbj-prn} is very competent .\\
When the \kw{\{f/m\}-occ-sg} arrived, everyone applauded \kw{\{f/m\}-obj-prn} .\\
The \kw{\{f/m\}-occ-pl} with the highest success rate per the polls, were \kw{\{f/m\}-n} .\\
That \kw{\{f/m\}-occ-sg} is also my \kw{\{f/m\}-rel} .\\
The \kw{\{f/m\}-occ-sg} moved here from Ethiopia where \kw{\{f/m\}-sbj-prn} is from .\\
The \kw{\{f/m\}-occ-sg} arrives early for \kw{\{f/m\}-pos-prn} shift .\\
That \kw{\{f/m\}-occ-sg} is hella strict, watch out for \kw{\{f/m\}-obj-prn} . \\
That \kw{\{f/m\}-occ-sg} retired early, good for \kw{\{f/m\}-obj-prn} . \\\bottomrule
\end{tabular}%
}
}
\caption{Our template set used to generate English source sentences. }
\label{tab:full-template}
\end{table*}

\subsection{Breakdown of scatter plots}\label{sec:breakdown}
\begin{figure}[b]
\centering
	\begin{subfigure}[t]{0.48\textwidth}
    \centering
  \includegraphics[width=2.5in]
  {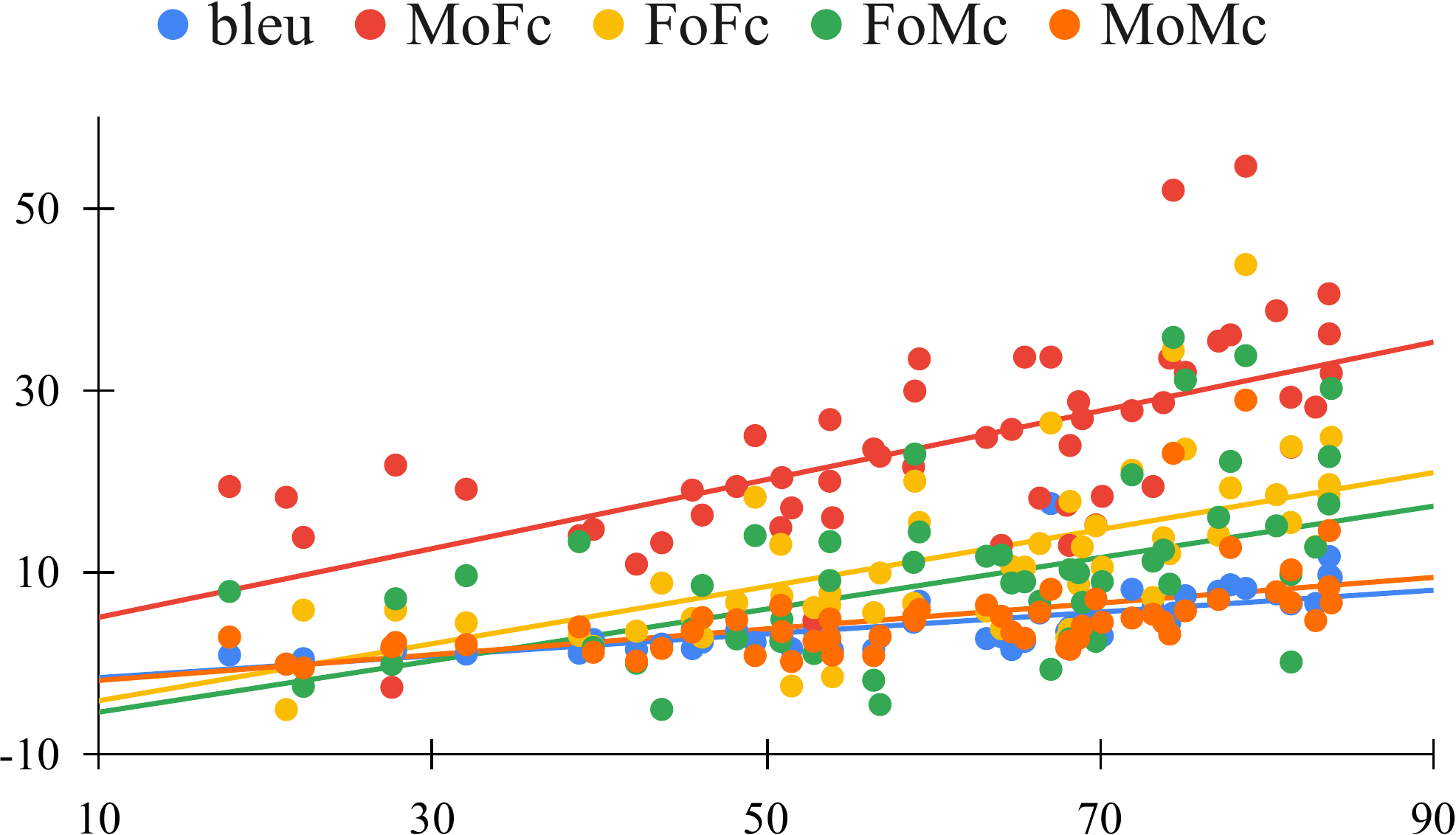}
		\caption{English-Spanish}
		\vspace{3mm}
		\label{fig:scatter-en-es}		
	\end{subfigure}	\quad
	\begin{subfigure}[t]{0.48\textwidth}
    \centering
  \includegraphics[width=2.5in]
  {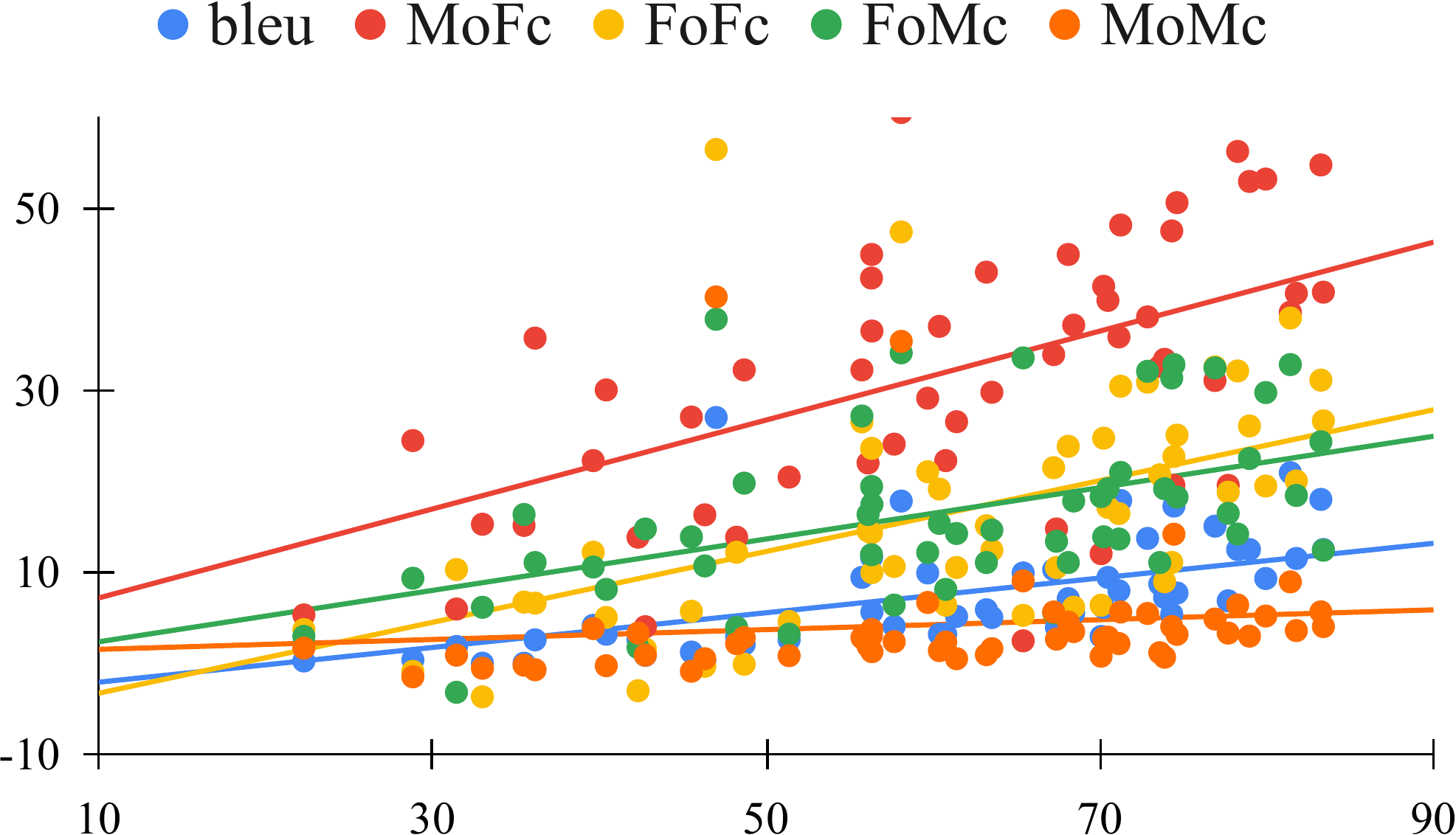}
		\caption{English-German}\label{fig:scatter-en-de}
	\end{subfigure}
	\caption{Plots showing \emph{relative percentage drop} of BLEU and gender-test metrics on the $y$-axis and \emph{relative percentage drop} in decoding time in the $x$-axis. 
	}
	\label{fig:scatter-breakdown}
\end{figure}

\Cref{fig:scatter-en-es,fig:scatter-en-de} further divides pro-stereotypical into male-occupations in male contexts (MoMc) and female-occupations in female context (FoFc), and anti-stereotypical into male-occupations in female contexts (MoFc) and female-occupations in male contexts (FoMc).

\subsection{Evaluation Dictionary}
\Cref{tab:dictionary} shows the dictionary we use for evaluation.
\begin{table*}[]
\centering
\resizebox{\textwidth}{!}{%
\begin{tabular}{@{}lllll@{}}
English & Spanish-Masc & Spanish-Fem & German-Masc & German-Fem \\ \midrule
clerk & funcionario\textbar  empleado administrativo & funcionaria\textbar  empleada administrativa & Mitarbeiter & Mitarbeiterin \\
designer & diseñador & diseñadora & Designer & Designerin \\
hairdresser & peluquero & peluquera & Friseur & Friseurin \\
housekeeper & amo de llaves\textbar  amo de casa & ama de llaves\textbar  ama de casa & Haushälter & Haushälterin \\
nanny & niñero & niñera & Kinderbetreuer & Kindermädchen \\
nurse & enfermero & enfermera & Krankenpfleger & Krankenschwester \\
secretary & secretario & secretaria & Sekretär & Sekretärin \\
director & director & directora & Direktor & Direktorin \\
engineer & ingeniero & ingeniera & Ingenieur & Ingenieurin \\
truck driver & camionero & camionera & Lastwagenfahrer & Lastwagenfahrerin \\
farmer & granjero\textbar  campesino & granjera\textbar  campesina & Bauer & Bäuerin \\
laborer & obrero\textbar  trabajador & obrera\textbar  trabajadora & Arbeiter & Arbeiterin \\
mechanic & mecánico & mecánica & Mechaniker & Mechanikerin \\
physician & médico & médica & Arzt & Ärztin \\
president & presidente & presidenta & Präsident & Präsidentin \\
plumber & plomero\textbar  fontanero & plomera\textbar fontanera & Klempner & Klempnerin \\
carpenter & carpintero & carpintera & Tischler\textbar Zimmermann & Tischlerin \\
groundskeeper & jardinero\textbar  guardián & jardinera\textbar  guardiana & Gärtner & Gärtnerin \\
clerks & funcionarios & funcionarias & MItarbeiter & Mitarbeiterinnen \\
designers & diseñadores & diseñadoras & Designer & Designerinnen \\
hairdressers & peluqueros & peluqueras & Friseure & Friseurinnen \\
housekeepers & amos de llaves\textbar  amos de casa & amas de llaves\textbar  amas de casa & Haushälter & Haushälterinnen \\
nannies & niñeros & niñeras & Kinderbetreuer & Kindermädchen \\
nurses & enfermeros & enfermeras & Krankenpfleger & Krankenschwestern \\
secretaries & secretarios & secretarias & Sekretäre & Sekretärinnen \\
directors & directores & directoras & Direktoren & Direktorinnen \\
engineers & ingenieros & ingenieras & Ingenieuren & Ingenieurinnin \\
truck drivers & camioneros & camioneras & Lastwagenfahrerin & Lastwagenfahrerinnen \\
farmers & granjeros & granjeras & Bauern & Bäuerinnen \\
laborers & obreros & obreras & Arbeiter & Arbeiterinnen \\
mechanics & mecánicas & mecánicos & Mechaniker & Mechanikerinnen \\
physicians & médico & médicas & Ärzte & Ärztinnen \\
presidents & presidentes & presidentas & Präsidenten & Präsidentinnen \\
plumbers & plomeros & plomeras & Klempner & Klempnerinnen \\
carpenters & carpinteros & carpinteras & Tischler & Tischlerinnen \\
groundskeepers & jardineros\textbar  guardianes & jardineras\textbar  guardianas & Gärtner & Gärtnerinnen \\ \bottomrule
\end{tabular}%
}
\caption{Our dictionary of occupations. Entries with the ``\textbar'' symbol indicate that we accept either of the references as correct.}
\label{tab:dictionary}
\end{table*}
\end{document}